\documentclass[11pt, a4paper, singlecolumn]{article}

\usepackage{color}

\usepackage[utf8]{inputenc}
\usepackage[english]{babel}

\usepackage[top=2cm, bottom=2cm, left=1.9cm, right=1.9cm]{geometry}

\usepackage{titlesec}
\titlespacing{\section}{0pt}{\parskip}{\parskip}
\titlespacing{\subsection}{0pt}{\parskip}{\parskip}
\titlespacing{\subsubsection}{0pt}{\parskip}{\parskip}

\usepackage{indentfirst}
\usepackage[export]{adjustbox}
\usepackage{graphicx}
\usepackage{amssymb}
\usepackage{amsmath}

\usepackage[dvipsnames]{xcolor}

\usepackage[figurename=Fig.]{caption}
\usepackage{caption}
\usepackage[colorlinks, citecolor=ForestGreen]{hyperref}

\usepackage{cite}

\usepackage{abstract}
\usepackage{multirow}
\usepackage{array}
\newcolumntype{*}{>{\global\let\currentrowstyle\relax}}
\newcolumntype{^}{>{\currentrowstyle}}

\usepackage{subcaption}

\title{\textbf{Using Motion History Images with 3D Convolutional Networks in Isolated Sign Language Recognition}}
\date{\vspace*{2pt}}
\author{
	\normalsize \textbf{Ozge Mercanoglu Sincan}\\
	\normalsize Ankara University\\
	\normalsize Computer Engineering Department\\
	\normalsize omercanoglu@ankara.edu.tr
	\and
	\normalsize \textbf{Hacer Yalim Keles}\\
	\normalsize Hacettepe University\\
	\normalsize Computer Engineering Department\\
	\normalsize hacerkeles@cs.hacettepe.edu.tr
}

\begin{document}
\maketitle

\begin{abstract}{
	\vspace*{-1.5em}
	\it 
	Sign language recognition using computational models is a challenging problem that requires simultaneous spatio-temporal modeling of the multiple sources, i.e. faces, hands, body, etc. In this paper, we propose an isolated sign language recognition model based on a model trained using Motion History Images (MHI) that are generated from RGB video frames. RGB-MHI images represent spatio-temporal summary of each sign video effectively in a single RGB image. We propose two different approaches using this RGB-MHI model. In the first approach, we use the RGB-MHI model as a motion-based spatial attention module integrated into a 3D-CNN architecture. In the second approach, we use RGB-MHI model features directly with the features of a 3D-CNN model using a late fusion technique. We perform extensive experiments on two recently released large-scale isolated sign language datasets, namely AUTSL and BosphorusSign22k. Our experiments show that our models, which use only RGB data, can compete with the state-of-the-art models in the literature that use multi-modal data.
	
}\end{abstract}

\textbf{Keywords} --- 3D-CNN, attention, deep learning, motion history image, sign language recognition.

\section{Introduction}
According to the World Federation of the Deaf (WFD), there are 70 million deaf people around the world \cite{wfd}. People in deaf communities use sign language for communication with each other and with hearing people. Sign languages are visual languages that use manual articulations of hands in combination with facial expression and body posture to represent signs. Each country has its own sign language, and sign languages have their own lexicon and grammar. The fact that the majority of hearing people do not know sign language poses a challenge for deaf communities in daily life. To overcome this challenge, computer vision researchers carry out studies on automatic Sign Language Recognition (SLR) systems.

In the literature, there are two main tracks of related research on the SLR domain; isolated and continuous SLR. While a video sample includes only one sign in an isolated SLR setting, continuous SLR videos contain multiple signs and require recognition of a sequence of multiple signs. Although continuous SLR is more challenging, isolated SLR is still an active research area. Particularly, the recently released new large-scale datasets provide unique challenges to the community \cite{joze2018ms, huang2018attention, li2020word, sincan2020autsl, ozdemir2020bosphorussign22k, hu2020global}. According to the survey in \cite{koller2020quantitative}, there is an exponential growth in isolated SLR studies between 1983-2020. 

Computationally, isolated SLR can be considered similar to an action recognition problem. However, SLR involves the discrimination of a sequence of fine-grained local motions that are usually performed quickly. For similarly performed signs, only small differences make signs different; for instance, for some pairs of signs, hand gestures look very similar, yet there is a difference in the facial expressions. For some pairs of signs, hand motion trajectories look very similar, yet local hand gestures look slightly different. For some pairs of signs, although hand gestures and trajectories are the same, the number of repetitions of a gesture is different. To deal with these challenges, most studies explicitly segment the hands or face regions in video frames, or use multiple modalities, such as depth, skeleton, etc. to obtain higher accuracies in sign classification. 

In this research, we aim to propose an effective solution, which uses only RGB images in the sign videos without requiring any explicit part segmentation or additional data modalities. We create RGB-Motion History Images (RGB-MHI) for each video and train an SLR model using only RGB-MHI images. Based on this model, we propose two different approaches that use RGB-MHI images and RGB frames of sign videos: In the first approach, we experiment with a 3D Convolutional Neural Network (3D-CNN) using our RGB-MHI model features as an attention mechanism to focus on relevant parts of the video frames. In the second approach, RGB frames and RGB-MHI images are used together with a late fusion technique. Our empirical observations show that using RGB-MHI images with 3D-CNN models provides comparable performances with the state-of-the-art models that use multiple input modalities, such as depth, skeleton, hand and face segmentation, etc., or multiple model ensembles. The proposed solutions in this research are considerably cheaper with respect to their memory and computation requirements.  

The remainder of the paper is organized as follows: In Section \ref{sec:RelatedWorks}, we summarize the related works. In Section \ref{sec:methods}, we describe our proposed methods. In Section \ref{sec:experiments}, we provide the experiments and model training details. Finally, we conclude the paper in Section \ref{sec:conclusion}.

\section{Related Works}
\label{sec:RelatedWorks}

Feature extraction is a crucial step for isolated sign language recognition as it is for other problems in computer vision. In the literature, some early studies utilized mainly hand features for sign language recognition by utilizing glove-based models, or using different devices, such as Leap Motion Controller (LMC), which aims to detect and track hands and fingers \cite{shukor2015new},\cite{katilmics2021elm}. Tracking only the hand region is not sufficient for general-purpose sign recognition. The signs in sign languages are identified by manual sources and non-manual sources simultaneously. While manual sources consist of 4 components: hand shape (finger configurations), orientation of the palm, movement of the hand, and position of the hand; non-manual sources include body posture and facial expressions such as head tilting, mouthing, etc. \cite{al2021deep}. A powerful SLR system should follow all these body parts simultaneously. Therefore, in most recent SLR studies, the methods approach the problem holistically; analyzing all components of the signer data as a whole.

The early research in the field use hand-crafted features such as HOG, Hu moments, motion velocity vector, relative hand positioning \cite{jangyodsuk2014sign, memics2013turkish, rao2018selfie} as features. After the release of the new large-scale datasets and advancements in deep neural networks, spatio-temporal sign features are successfully learned from the data distribution. Convolutional deep models are useful in this regard to learn spatial features automatically from data \cite{pigou2018beyond, sincan2019isolated, luqman2021towards}. In these studies, after spatial features are extracted by 2D-CNNs, temporal information is modelled by a recurrent neural network. On the other hand, some studies use the combination of deep learning and traditional methods. Rastgoo \emph{et al.} \cite{rastgoo2020video} used some hand-crafted features and 2D-CNNs to obtain spatial information. Then, they fuse all the features and feed them to LSTM for temporal feature extraction. 

Recently, 3D-CNNs show remarkable performances in modeling the spatio-temporal patterns simultaneously in video frames. Joze \emph{et al.} \cite{joze2018ms} released a large-scale American Sign Language (ASL) dataset, namely MS-ASL, and provided baseline methods including 2D-CNN-LSTM and 3D-CNN based models. A 3D-CNN based I3D model \cite{carreira2017quo} achieved the best result with a large improvement to the 2D-CNN based model. Li \emph{et al.} \cite{li2020word} released another large-scale isolated ASL dataset namely WLASL, and they experimented with several appearance-based and pose-based deep learning methods; (a) 2D-CNN network and Gated Recurrent Unit (GRU), (b) 3D-CNN network, (c) human pose-based GRU (d) human pose-based temporal graph convolution network. Their results also show that 3D-CNN based I3D achieved the best results. Deep learning architectures that work with a sequence of images require considerable processing power and memory requirements. Therefore, some studies work on more efficient representations; Dos Santos \emph{et al.} \cite{dos2020dynamic} created colored motion history images (MHI), namely the star RGBs, to represent video sequences. They trained two ResNets and combine these models with a soft-attention mechanism. They achieved significant accuracy rates in three different public datasets using only RGB data. Imran and Raman \cite{imran2020deep} also represented a video in a single image. They created three different kinds of motion templates: MHI, RGB motion image, dynamic image. In our proposed approach, we use a similar single MHI image representation to \cite{dos2020dynamic}, yet we compute MHI images as in \cite{barros2014real}. 

In SLR research, it is a common approach to use different types of data modalities such as depth, skeleton, optical flow, etc. in addition to RGB data, and combine these multiple modalities with a fusion technique in order to boost the accuracy rate. Jiang \emph{et al.} \cite{jiang2021skeleton}, the winner team of the ChaLearn 2021 Looking at People Large Scale Signer Independent Isolated SLR CVPR Challenge \cite{sincan2021chalearn}, utilized different types of data modalities, e.g., RGB, skeleton, optical flow, depth, depth HHA, depth flow. They mention that multi-modal ensembles obtain a higher accuracy rate than every single modality. Some studies try to segment hand or face regions to use them as an additional modality. Luqman and El-Alfy \cite{luqman2021towards} utilized Microsoft Kinect for this purpose. Firstly, they trained several 2D-CNN-LSTM models using color, depth, or optical flow data separately. Then, they used animation units of the signer face provided by the Kinect as input and trained a stacked LSTM model. They fused these manual and non-manual features at the classification level with their best 2D-CNN-LSTM model and improved the accuracy rate by about 3\%. Some studies try to segment these regions from RGB data without using any hardware but with some extra preprocessing. Using a pose estimation algorithm, such as OpenPose \cite{cao2019openpose}, is one of the popular methods to detect body, face, and hand keypoints. In \cite{gruber2021mutual} and \cite{de2021isolated}, hand regions were cropped by creating a bounding box around the hand keypoints obtained using OpenPose. They trained several models separately using different types of data including cropped hands and proposed an ensemble model in order to increase the recognition accuracy. Gökçe \emph{et al.} \cite{gokce2020score} also utilized OpenPose to crop both face and hand regions, and used these modalities together with the full-body images. Rastgoo \emph{et al.} \cite{rastgoo2018multi} used Faster R-CNN model \cite{ren2016faster} to detect hand regions, and they created three forms of input; original image, cropped hand image, and noisy cropped hand image. They proposed a Restricted Boltzmann Machine (RBM) using RGB and depth modalities to perform automatic hand sign language recognition.

In parallel to these, using attention mechanisms to focus on relevant spatial regions or temporal video frames without needing explicit segmentation have been also studied in the SLR domain. Shi \emph{et al.} \cite{shi2019fingerspelling} proposed an iterative visual attention mechanism for ASL fingerspelling. They aimed to focus on signing hand with an attention model, which is based on a convolutional recurrent architecture. Their attention model reduced the area of interested regions iteratively, and while doing this, increased the resolution. Huang \emph{et al.} \cite{huang2018attention} proposed an attention-based 3D-CNN for isolated SLR. In their proposed model, they extracted spatio-temporal features with a C3D model. A spatial attention mechanism was incorporated into C3D by using skeleton information of hands and arms. Then, they built a temporal attention-based model for classification. In our previous work \cite{sincan2020autsl}, we proposed a baseline model for a new large-scale isolated Turkish Sign Language (AUTSL) dataset. We integrated a Feature Pooling Module and a temporal attention model to focus on more relevant spatio-temporal parts of the videos. Hu \emph{et al.} \cite{hu2020global} collected non-manual features aware isolated CSL dataset, which contains visually similar confusing signs that only differ in their non-manual features. They proposed 3D-Resnet-50 based \cite{qiu2017learning} model that consists of two cooperative global and local enhancement modules to differentiate confusing signs. Global enhancement module contains a self-attention mechanism based on a Non-local Neural Network \cite{wang2018non}, with the aim of enhancing the global contextual relationship. In our proposed work, we utilize the RGB-MHI images as an attention model to assist our 3D-CNN model to focus on spatially salient regions.

\section{Proposed Method}
\label{sec:methods}
In this paper, we aim to construct a sign language recognition model using only RGB data, without using any explicit part segmentation of hands or face regions. In this regard, we create an RGB-MHI image that summarizes the entire video in a single video frame and propose a RGB-MHI model that learns a representation of relevant spatial and motion patterns using these single images. Then, we propose two different approaches utilizing this model. In the first approach, we use the RGB-MHI model as a motion-based attention mechanism to focus on relevant spatial regions. In the second one, we propose a fusion model that combines RGB and RGB-MHI features.

\begin{figure*} 
	\centering    
	\includegraphics[width=0.8\textwidth]{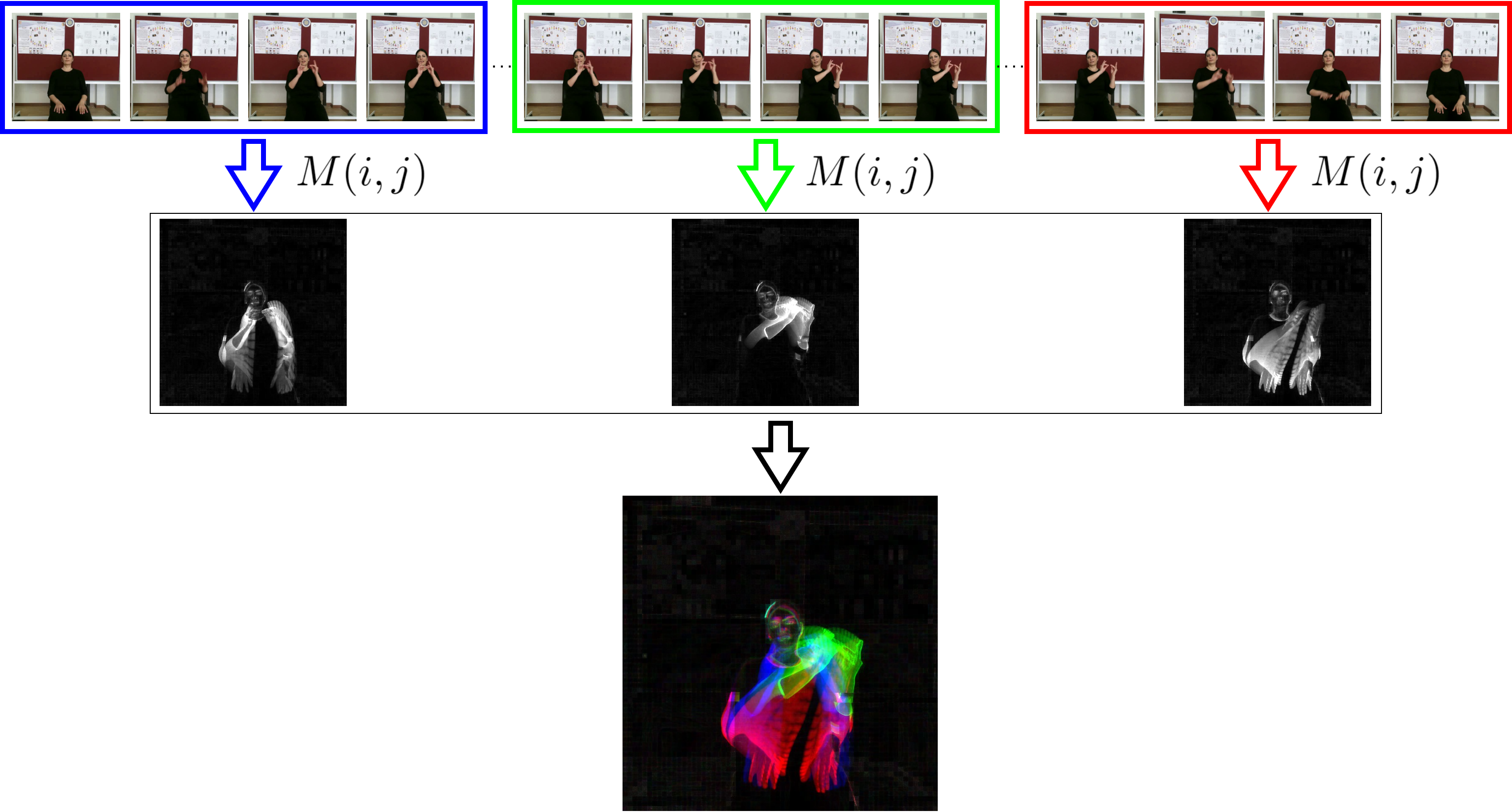}
	\caption{An example of creating an RGB-MHI image.}
	\label{fig:rgb-mhi-example}
\end{figure*}

\subsection{RGB-Motion History Image Model}
\label{sec:mhi}
Motion History Image (MHI) \cite{bobick2001recognition} is a static gray-scale image  where more recently moving pixels are brighter; there are different variants of MHI image generation \cite{AlpKeles2017}, \cite{barros2014real}. In \cite{barros2014real}, MHI is estimated by the sum of the absolute values of differences between consecutive frames as in \eqref{eq:mhi1}, where $N$ is the number of frames in a video; $I\textsubscript{t}$ is the $t\textsuperscript{th}$ video frame; $(i,j)$ are the coordinates of a pixel in a frame and $W$ represents the weight for the absolute difference, which is calculated as $W = t/N$. In \cite{dos2020dynamic}, a modified version of (\ref{eq:mhi1}) is proposed by including magnitude and phase information as well. They also propose an approach to represent MHIs as RGB images. In this approach, the frames in a video stream are split into three equal parts, and motion histories are calculated for each part separately. Then, a 3-channel colored MHI image, which is referred to as star RGB, is created. In this representation B-channel contains the motion history information from the first temporal region, the G-channel has the motion history information from the central one, and the R-channel from the last part. Dividing the video frames into 3 parts helps prevent temporal information loss.

\begin{equation}
	\label{eq:mhi1}
	M(i,j) = \sum_{t=2}^{N} | I\textsubscript{t-1}(i,j) - I\textsubscript{t}(i,j) | W
\end{equation}

In our proposed method, we also consider using colored MHI images for SLR as in \cite{dos2020dynamic}, yet while generating our RGB-MHI images we use the differences between two consecutive frames as in \eqref{eq:mhi1} for its simplicity. We depicted our colored RGB-MHI image generation approach in Fig. \ref{fig:rgb-mhi-example}.

Our RGB-MHI model is based on ResNet-50 \cite{he2016deep}, which achieved successful results with star RGB in gesture recognition problem \cite{dos2020dynamic}. We use pre-trained ResNet-50 model on ImageNet \cite{deng2009imagenet} dataset, and fine-tune it to the used sign language datasets. However, RGB-MHI images are contextually different from normal images. Therefore, we made some ablation studies to select ResNet layers depending on their performances; we wanted to reduce the number of layers without sacrificing the model performance. However, we observed that the classification accuracies were lower than we expected when we used the Global Average Pooling (GAP) as it was used in the original ResNet model. We believe that losing the details in the spatial context affects model performances considerably with MHI images. Therefore, we removed the GAP layer and included an additional fully connected (FC) layer with 1024 neurons, with a ReLU non-linearity; the last FC layer follows this layer with a softmax function. We obtained the best results when we included all the convolutional layers in the pre-trained ResNet-50 model. The details of the experimental results are provided in Section \ref{sec:ablation}. As we expected, 2D-CNN models converge much faster when we use single RGB-MHI images than using a sequence of RGB video frames.

\subsection{3D-CNN Models}
\label{sec:3dcnnModels}

3D-CNN models have recently been widely used in video classification problems in different computer vision domains and provide state-of-the-art results. Motivated from this, in this study, we planned preliminary experiments with several successful 3D-CNN models to select the base 3D model that we can use together with our RGB-MHI model. In this context, we first worked with the C3D \cite{tran2015learning} model that is pre-trained on Sports-1M \cite{karpathy2014large} dataset. Secondly, we experimented with the Inflated 3D ConvNet (I3D) \cite{carreira2017quo} model that is pre-trained on Kinetics-400 \cite{kay2017kinetics} dataset. With the original C3D model, which has 8 convolutional layers, 5 pooling layers, 2  FC layers, we had a vanishing gradient problem. Therefore, we made some modifications in the architecture; we replaced the 2 FC layers with a global average pooling, and also we inserted a batch normalization (BN) layer after all the convolutional layers. Compared to the C3D model, the I3D model has fewer parameters and achieved better results; recently, it has been used on new sign language datasets, e.g., WLASL\cite{li2020word}, MS-ASL \cite{joze2018ms}, and BSL-1K \cite{albanie2020bsl}. In \cite{albanie2020bsl}, it was observed that I3D achieved higher accuracy on two ASL datasets when pre-trained on a new large-scale British Sign Language dataset, BSL-1K, compared to Kinetics-400. Therefore, we experimented with an I3D model that is pre-trained using both Kinetics-400 and BSL-1K datasets.

In recent years, for efficiency reasons, 3D convolutions are separated into a 2D spatial convolution and a 1D temporal convolution. In these methods, 3D-CNNs are replaced with spatial and temporal separable 3D convolutions, i.e., t × k × k is replaced with 1 × k × k followed by t × 1 × 1, where t is the width of the filter in time, and k is the width of the filter in spatial extent \cite{xie2018rethinking, tran2018closer}. In this work, we also experimented with two (2+1)D models, Separable 3D-CNN (S3D) \cite{xie2018rethinking} and R(2+1)D-18 \cite{tran2018closer}.  S3D is the separable version of the I3D architecture and has much fewer parameters than the original I3D. On the other hand, although R(2+1)D does not reduce the number of parameters much, it eases the optimization process.

For all the architectures, we used publicly available pre-trained models and fine-tune them to the used sign language datasets. At the end of the preliminary experiments, we decided to work with the I3D model \cite{carreira2017quo} that is pre-trained on BSL-1K \cite{albanie2020bsl}. 

\subsection{3D-CNN Models with Attention Mechanism}

Attention mechanisms have been included in deep models in various ways to improve the performances of different tasks in computer vision and natural language processing domains. It has also been used in video classification problems, such as action recognition and sign language recognition, to focus on more relevant spatial or temporal parts of the stream and construct more robust models. In this study, we propose to utilize our RGB-MHI model as an attention model with the I3D model to focus on the spatial regions where there is a motion cue. In addition to our RGB-MHI based attention model, we also implemented a separate self-attention integrated I3D model. In this section, we first present the details of our self-attention integrated I3D model, then we describe our proposed motion-based attention mechanism.

\subsubsection{Self-attention Mechanism}
\label{sec:self-attentionModel}

The convolutional operators in CNNs, process a region, i.e. computes the responses of filters, considering the values in their local neighborhood; hence, long-range dependencies are not captured. A self-attention mechanism aims to learn long-range dependencies \cite{vaswani2017attention}. The embedded Gaussian version of Non-Local (NL) blocks \cite{wang2018non} is a kind of self-attention that is proposed for computer vision tasks. In this architecture, a non-local operation computes a response for an output position by the weighted sum of the features from all positions. 

Motivated from \cite{wang2018non}, we incorporate NL blocks to some of the middle-layers of I3D architecture. In this context, we feed a feature map, $X \in R^{C\text{x}T\text{x}H\text{x}W}$, into three different convolutional layers $(\theta, \phi, g)$ with filters of size 1 × 1 × 1, where $C, T, H, W$ represent the channel, temporal, height, and width dimensions, respectively. The NL operation is described as in \eqref{eq:nloperation}, where $i$ is the index of output position, $j$ is the index of all possible positions, $X$ is the input feature map, and $x$ is the output feature map. With all convolutional layers in (\ref{eq:nloperation}), we reduce the channel dimension to half, thus the new channel dimension becomes $C/2$. Also, we apply a max pooling layer after  $\phi$ and $g$ to reduce the amount of pairwise computation. Then, the obtained feature map, $x\textsubscript{i}$, is fed into another 1 × 1 × 1 convolutional layer with $C$ channel dimension, and the result is added to initial feature map $X\textsubscript{i}$ as in \eqref{eq:nlblock}.

\begin{equation}
	x\textsubscript{i} = softmax(\theta(X\textsubscript{i})\textsuperscript{T}\phi(X\textsubscript{j}))g(X\textsubscript{j})
	\label{eq:nloperation}
\end{equation}

\begin{equation}
	z\textsubscript{i} = W\textsubscript{z}x\textsubscript{i} + X\textsubscript{i} 
	\label{eq:nlblock}
\end{equation}

We conduct some experiments by adding a various number of NL blocks to different middle-layers. We achieve the best accuracy rate when we add 1 NL block after the third block (3a) of the I3D model. Our experimental results are given in the Table \ref{tab:attention}.

\subsubsection{The Proposed Motion-based Attention Mechanism}
\label{sec:attentionModel}

In order to assess the salient regions using our pre-trained RGB-MHI model, we first apply channel-based global average pooling to the feature maps at the output of one of the middle layers. Then, we create an attention weight by passing the obtained feature matrix through softmax function and then normalizing the result between $(0,1)$ as follows:

\begin{equation}
	\alpha = softmax(\sum\limits_{j=1}^{W} \sum\limits_{i=1}^{H} \frac{1}{C} \sum\limits_{c=1}^{C} Y\textsubscript{c,i,j})
	\label{eq:motionAttention}
\end{equation}

\begin{equation}
	\alpha_{norm} = \frac{\alpha - min(\alpha)}{max(\alpha)-min(\alpha)} 
	\label{eq:normalize}
\end{equation}

where $Y\in R^{C\text{x}H\text{x}W}$ is the feature map of a middle-layer in RGB-MHI model; $\alpha \in R^{H\text{x}W}$; $C, H, W$ represent the channel, height, and width dimensions, respectively.
The normalized attention weights, $\alpha_{norm}$, indicate the salient regions in a video. As ablation studies, we generated attention weights from three different layers of the RGB-MHI model, i.e. conv\textsubscript{2}, conv\textsubscript{3}, or conv\textsubscript{4}.  After the attention weights are obtained, it is element-wise multiplied with the feature maps in the I3D model, which has the same spatial size with $\alpha_{norm}$; i.e. 2a, 3a, or 4a layers of the I3D. 

Our proposed method is illustrated in Fig. \ref{fig:methods}a. In the Figure, we depict the motion-based attention aggregation in 3\textsuperscript{rd} layer, since we observed the best accuracy rates using this layer.

\subsection{Fusion of I3D and RGB-MHI Models}
\label{sec:ensemble}

\begin{figure*}  [htbp] 
	\centering    
	\includegraphics[width=0.99\textwidth]{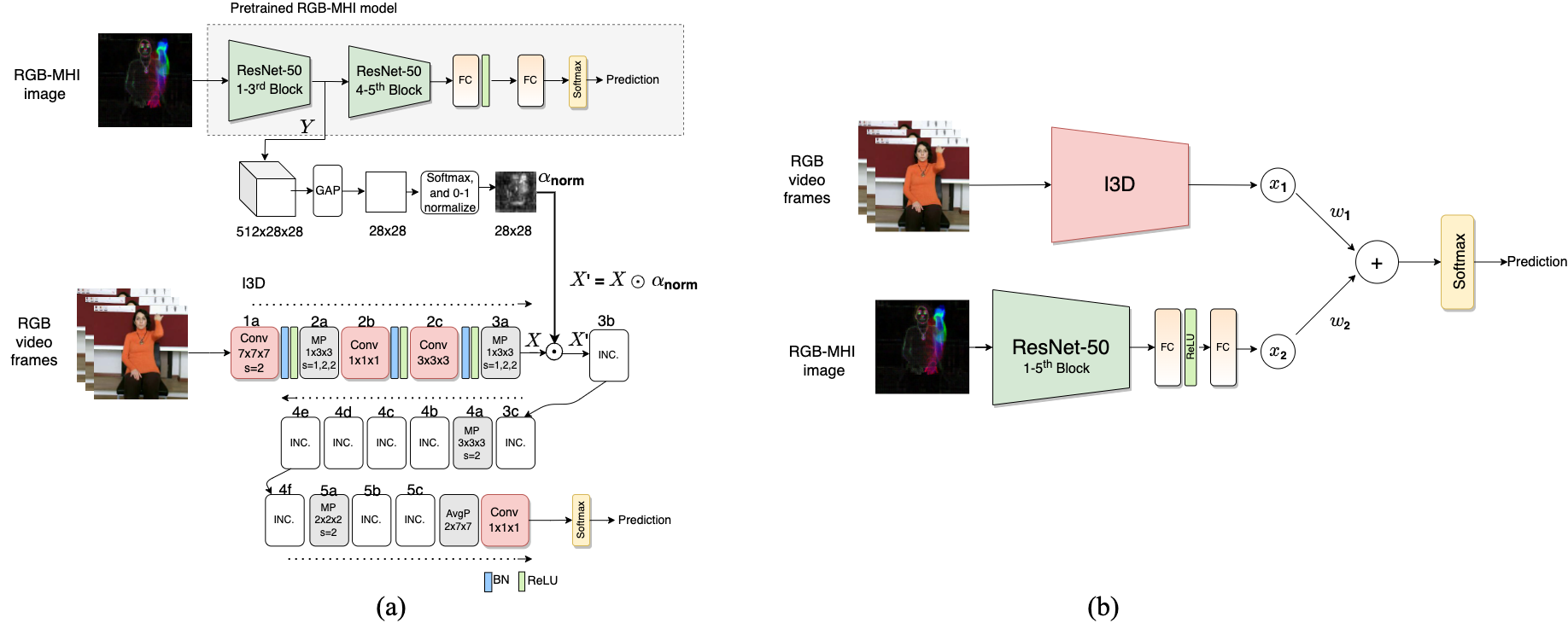}
\caption{Proposed methods. (a) I3D with RGB-MHI attention (b) I3D+RGB-MHI fusion.}
	\label{fig:methods}
\end{figure*}

In the literature, some studies show that fusing multi-modalities or multi-model ensembles achieve higher accuracies than a single model. Therefore, we fuse our I3D and RGB-MHI models with a late fusion technique. In this approach, we utilize separate versions of I3D and RGB-MHI, without parameter updating. We assign $w\textsubscript{1}$ and $w\textsubscript{2}$ weights for the output of the last layers before softmax, and then calculate their weighted sum as in \eqref{eq:prediction} for the final prediction.  Our proposed fusion method is illustrated in Fig. \ref{fig:methods}b.

\begin{equation}
	prediction = softmax(w\textsubscript{1}x\textsubscript{1} + w\textsubscript{2}x\textsubscript{2})
	\label{eq:prediction}
\end{equation}

\section{Experiments}
\label{sec:experiments}
This section describes the datasets, training details, various ablation studies, and comparisons to the state-of-the-art model performances in the literature.

\subsection{Datasets and Preprocessing}
\label{sec:datasets}
We evaluate our proposed models on two very recently shared large-scale isolated sign language datasets: AUTSL \cite{sincan2020autsl} and BosphorusSign22k \cite{ozdemir2020bosphorussign22k}.

\textbf{AUTSL} \cite{sincan2020autsl} is a large-scale, signer independent, isolated TSL dataset that contains 226 signs, and 36,302 video samples. The dataset contains 43 signers; 31 of them are included in the training set, 6 are in the validation set, and the remaining 6 are in the test set. This benchmark provides a signer independent evaluation of the models. The methods are also challenged with 20 different backgrounds with dynamic elements, e.g., moving trees or moving people behind the signer. 

\textbf{BosphorusSign22k} \cite{ozdemir2020bosphorussign22k} is another large-scale, isolated TSL dataset that contains 744 signs, 22,542 video samples in which signs belong to health and finance domains, and also cover frequently used signs in daily activities. The dataset contains 6 signers; 1 of them is reserved for testing. In our experiments, we use 1 signer for the validation set. All signers are deaf people in this dataset.

\begin{figure} 
	\centering    
	\includegraphics[width=0.45\textwidth]{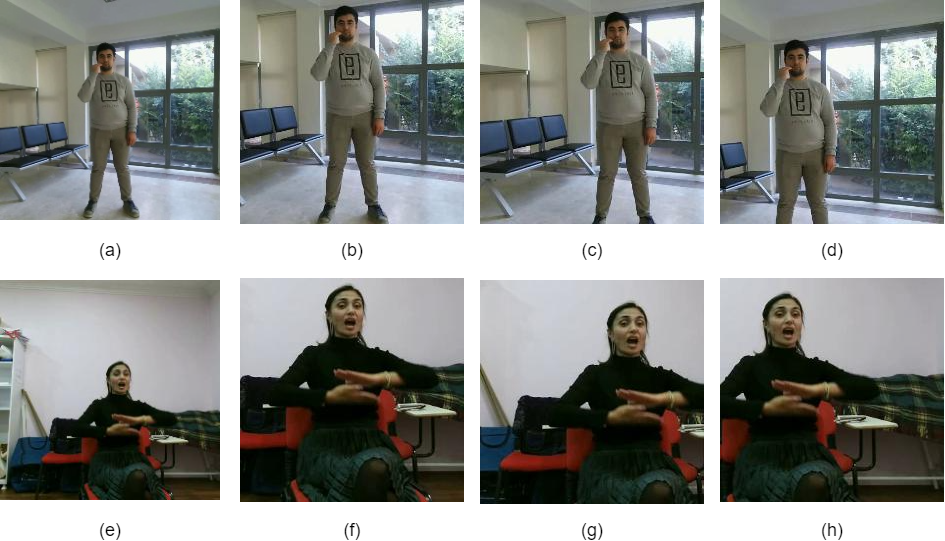}
	\caption{Examples of frame cropping and data augmentation of one standing, and one sitting signer from AUTSL dataset. (a, e) Original video frame. (b, f) After getting the bounding box with the Faster R-CNN \cite{ren2016faster}, the bounding box is extended equally from the left and right to be a square. (c, g) The bounding box is expanded from the left side first. (d, h) The bounding box is expanded from the right side first. While expanding from the right and left, the bounding box is randomly shifted in the vertical direction.}
	\label{fig:dataAug}
\end{figure}

\textbf{Data Preprocessing}: Firstly, we apply pre-trained Faster R-CNN model \cite{ren2016faster} with ResNet-50 Feature Pyramid Network (FPN) backbone to the first frames of the videos in order to detect signer and eliminate some of the irrelevant background details. In some videos of AUTSL, some people are passing by behind the signer. Therefore, when there is more than one person in the video we choose the largest person bounding box since the signer is always the closest person to the camera in this dataset and has the largest bounding box among people. We crop all video frames to be square by expanding the bounding box on both left and right sides (Fig. \ref{fig:dataAug}b, \ref{fig:dataAug}f). For 3D-CNN models, we fix the number of frames in all the videos to 32 frames. In isolated videos, the signer initially starts with a neutral position and returns to this position after performing the sign. For this reason, we skip some frames (maximum 10), which we determined according to the total number of frames in a video, from the beginning and the end of the stream. Then, we select 32 video frames from the remaining middle part by uniform sampling. On the other hand, while creating RGB-MHI images, we use all the video frames. After generating an RGB-MHI image, we crop it to be square using the bounding box information returned from Faster R-CNN of the corresponding video.

Moreover, in order to increase the variability, we apply data augmentation to RGB videos in the training and validation sets. In order to prevent the signers appear always in the center of the videos, we expand the bounding box from the left (Fig. \ref{fig:dataAug}c, \ref{fig:dataAug}g) or right (Fig. \ref{fig:dataAug}d, \ref{fig:dataAug}h), at first. In doing so, we also vertically shift the bounding box randomly (within a limited pixel) to allow the signer to be in different vertical positions. Finally, we also apply horizontal flip to these 3 versions of the data to accommodate our model for both hands. We resize all frames to 224x224.

\subsection{Training Details}
\label{sec:trainingdetails}

We implement all the models using PyTorch library \cite{paszke2019pytorch}. In our models, some experiments are conducted in order to determine the optimum hyperparameters. The best results are obtained as follows. In the RGB-MHI model, modified C3D, and R(2+1)D models, we use Adam optimizer with an initial learning rate $1e-4$. In I3D-based models, we use SGD optimizer with an initial learning rate $1e-2$, and momentum 0.9. In S3D, we use Adam optimizer with an initial learning rate $1e-3$. We reduce the learning rate by 0.2 factor if no improvement is observed on the validation set for three epochs in all models. We repeat this process several times, and when there is no improvement anymore, we finally terminate the training.

\subsection{Experiments on AUTSL}
\label{sec:ablation}

The results of our RGB-MHI model, which uses only one RGB-MHI image per video as an input, are reported in Table \ref{tab:rgb-mhi}. 
In this approach, we modified the ResNet-50 model and made empirical observations on several versions with different depths.  We achieved the best results when we used all the layers of ResNet-50. Compared to the baseline method, which is presented with the AUTSL dataset in \cite{sincan2020autsl}, our best RGB-MHI model achieves 71.95\% classification accuracy using a single image, which is higher than 2D-CNN and attention-based bidirectional LSTM model that takes entire video frames. Also, RGB-MHI model training is very fast since it only needs 1 motion history image, compared to 2D models where entire video frames are used as inputs. Moreover, since there are more than 200 different sign classes in the AUTSL dataset, 71.95\% accuracy rate shows that a single RGB-MHI is capable of learning important discriminative features to represent isolated signs.

\begin{table*}[htbp]
	\caption{Ablation studies with RGB-MHI model on AUTSL dataset. (The related feature sizes are shown in parentheses. Conv\textsubscript{1-5}: Block names in ResNet-50, Conv\textsubscript{extra}: Additional convolutional layer that we inserted, MP: Max Pooling, FC: Fully connected layer.)}
	\centering
    \small
    \renewcommand{\arraystretch}{1.2} 
	\label{tab:rgb-mhi}
	\begin{tabular}{ lc }
		\hline
		\textbf{Method} &  \textbf{Test Acc (\%)} \\ \hline
		Conv\textsubscript{1} (64x112x112), MP (64x56x56), FC (1024) & 38.52 \\ 
		Conv\textsubscript{2} (256x56x56), Conv\textsubscript{extra} (64x56x56), FC (1024) &  45.90 \\ 
		Conv\textsubscript{3} (512x28x28), Conv\textsubscript{extra} (128x28x28),  FC (1024) & 60.04 \\ 
		Conv\textsubscript{4} (1024x14x14), FC (1024) & 66.60 \\ 
		Conv\textsubscript{4} (1024x14x14), Conv\textsubscript{extra} (256x14x14), FC (1024)  & 68.77 \\ 
		\textbf{Conv\textsubscript{5} (2048x7x7), FC (1024)} & \textbf{71.95} \\ 
		Conv\textsubscript{5} (2048x7x7), Conv\textsubscript{extra} (512x7x7), FC (1024)  & 69.85 \\ \hline
		
	\end{tabular}
\end{table*}

Since there is still a large room for performance improvement, we experiment with 3D-CNN based models using only RGB video frames. The results of the ablation studies with 3D-CNN architectures on AUTSL are provided in Table \ref{tab:3dcnns}. Firstly, we experiment with the modified C3D model and achieve 83.04\% accuracy. Then, we experiment with the I3D model that recently achieves successful performances on the recent large-scale sign language datasets, as well as in action recognition. In our first experiment, the I3D model is pre-trained on Kinetics-400 \cite{kay2017kinetics} action recognition dataset, and the second one is pre-trained on BSL-1K \cite{albanie2020bsl} British sign language dataset. We observe that when we transfer the features from BSL-1K, we obtain 1.58\% higher accuracy. I3D that is pre-trained on BSL-1K, R(2+1)D-18, and S3D models all achieve above 89\% recognition accuracy. We choose the I3D model as the baseline to use in our follow-up experiments since many recent state-of-the-art works in SLR also chose I3D as their baseline, e.g., WLASL\cite{li2020word} and MS-ASL\cite{joze2018ms}.

In the follow-up works, we include the self-attention mechanisms to the selected 3D baseline model, i.e. I3D, to focus on related parts of the video frames. For this purpose, we insert different number of NL blocks \cite{wang2018non} after different stages motivated from the original paper, specifically after 3\textsuperscript{rd}, 4\textsuperscript{th}, or 5\textsuperscript{th} layers of I3D. In our experiments, we observe similar accuracy rates, but we get the best result (89.94\%) when we insert 1 NL block after 3a block of I3D, only a slight improvement (0.64\%) over the base model as seen in Table \ref{tab:attention}. Due to the memory constraints, we were able to add fewer NL blocks at lower level layers. However, we still obtain the best result when we add 1 NL block after the 3\textsuperscript{rd} layer. We believe that the 3\textsuperscript{rd} layer's larger spatial size is the main reason for that since it provides more spatial information.

\begin{table}[htbp]
	\caption{Ablation studies with 3D-CNNs on AUTSL dataset.}
	\centering
	\small
	\setlength\tabcolsep{4pt}
	\renewcommand{\arraystretch}{1.2} 
	\label{tab:3dcnns}
	\begin{tabular}{ lccc }
		\hline
		\textbf{Method} & \textbf{Params} & \textbf{Pretraining Dataset} &  \textbf{Test Acc (\%)} \\ \hline
		Modified C3D  & 27.7M & Sports1M \cite{karpathy2014large}  & 83.04 \\ 
		I3D  & 12.5M & Kinetics400 \cite{kay2017kinetics}  & 87.72 \\ 
		I3D  & 12.5M & BSL-1K \cite{albanie2020bsl}  & \textbf{89.30} \\ 
		R(2+1)D-18  & 31.4M & Kinetics400 \cite{kay2017kinetics}  & \textbf{89.67} \\
		S3D  &  8.1M & Kinetics400 \cite{kay2017kinetics} & \textbf{89.73}  \\ \hline
		
	\end{tabular}
\end{table}

\begin{table*}[htbp]
	\caption{Accuracy results of the ablation studies on self-attention and RGB-MHI attention on AUTSL dataset.}
	\centering
    \small
	\label{tab:attention}
	\begin{tabular}{ lc }
		\hline
		\textbf{Method} &  \textbf{Test Acc (\%)} \\ \hline
		I3D    & 89.30 \\ \hline
		\textbf{Self-attention:} &   \\ 
		I3D + 1 NL block after 3a & \textbf{89.94} \\ 
		I3D + 2 NL blocks after 4a, 5a  &89.38 \\ 
		I3D + 6 NL blocks after 4a, 4b, 4c, 4d, 4e, 4f  &  89.27 \\ 
	    I3D + 9 NL blocks after 4a, 4b, 4c, 4d, 4e, 4f, 5a, 5b, 5c & 89.57 \\  \hline
	    \textbf{RGB-MHI attention:} &   \\ 
	    I3D with RGB-MHI attention after 2a  & 89.38 \\ 
		I3D with RGB-MHI attention after 3a  & \textbf{90.18} \\ 
		I3D with RGB-MHI attention after 4a & 89.14 \\   \hline
		
	\end{tabular}
\end{table*}

\textbf{I3D Model with RGB-MHI Attention:} We conduct some experiments with our proposed motion-based attention model. In this scope, we create attention weights from the feature maps taken from one of the middle-layer, i.e. 2\textsuperscript{nd} or 3\textsuperscript{rd} or 4\textsuperscript{th} layer of RGB-MHI model. The sizes of the attention weights are 56×56, 28×28, 14×14, respectively. As in the experiments with the self-attention model, the best results in these experiments are obtained after the 3\textsuperscript{rd} layer, that is, when the size of attention weights are 28×28. Our RGB-MHI based attention model performs 90.18\% accuracy rate (Table \ref{tab:attention}), contributing 0.88\% to the basic I3D model and 0.24\% to the I3D model with self-attention.

\textbf{I3D and RGB-MHI Fusion Model:} 
In our second approach, we fuse I3D and RGB-MHI models with a late fusion technique and obtain 91.13\% with the weights of $w\textsubscript{1}$=0,6; $w\textsubscript{2}$=0,4. The empirical results show that RGB and RGB-MHI features are complementary to each other, thus the fusion of I3D and RGB-MHI model achieve better results than I3D with self-attention or I3D with RGB-MHI attention. Table \ref{tab:our-model} compares our proposed method variations.

\begin{table}[htbp]
	\caption{The results of our proposed method variations.}
	\centering
	\small
	\renewcommand{\arraystretch}{1.2} 
	\label{tab:our-model}
	\begin{tabular}{ lc }
		\hline
		\textbf{Method} &  \textbf{Test Acc (\%)} \\ \hline
		\textbf{Without data augmentation:} &   \\ 
		I3D & 89.30 \\
		I3D with self attention &  89.94 \\
		I3D with RGB-MHI attention &  90.18 \\ 
		I3D+RGB-MHI fusion  &  \textbf{91.13} \\ 
		I3D with RGB-MHI attention + fusion  &  \textbf{91.55} \\ \hline 
		\textbf{With data augmentation (x6 more data):} &   \\ 
		I3D & 92.19 \\ 
		I3D fine tune with train+validation data  & 92.43 \\ 
		I3D+RGB-MHI fusion, $(w\textsubscript{1}, w\textsubscript{2})$ = (0.7, 0.3) & 93.50 \\ 
		I3D+RGB-MHI fusion, $(w\textsubscript{1}, w\textsubscript{2})$ = (0.6, 0.4) & \textbf{93.53} \\ 
		I3D+RGB-MHI fusion, $(w\textsubscript{1}, w\textsubscript{2})$ = (0.5, 0.5) & 93.18 \\ \hline
	\end{tabular}
\end{table}

\begin{figure} 
	\centering    
	\includegraphics[width=0.48\textwidth]{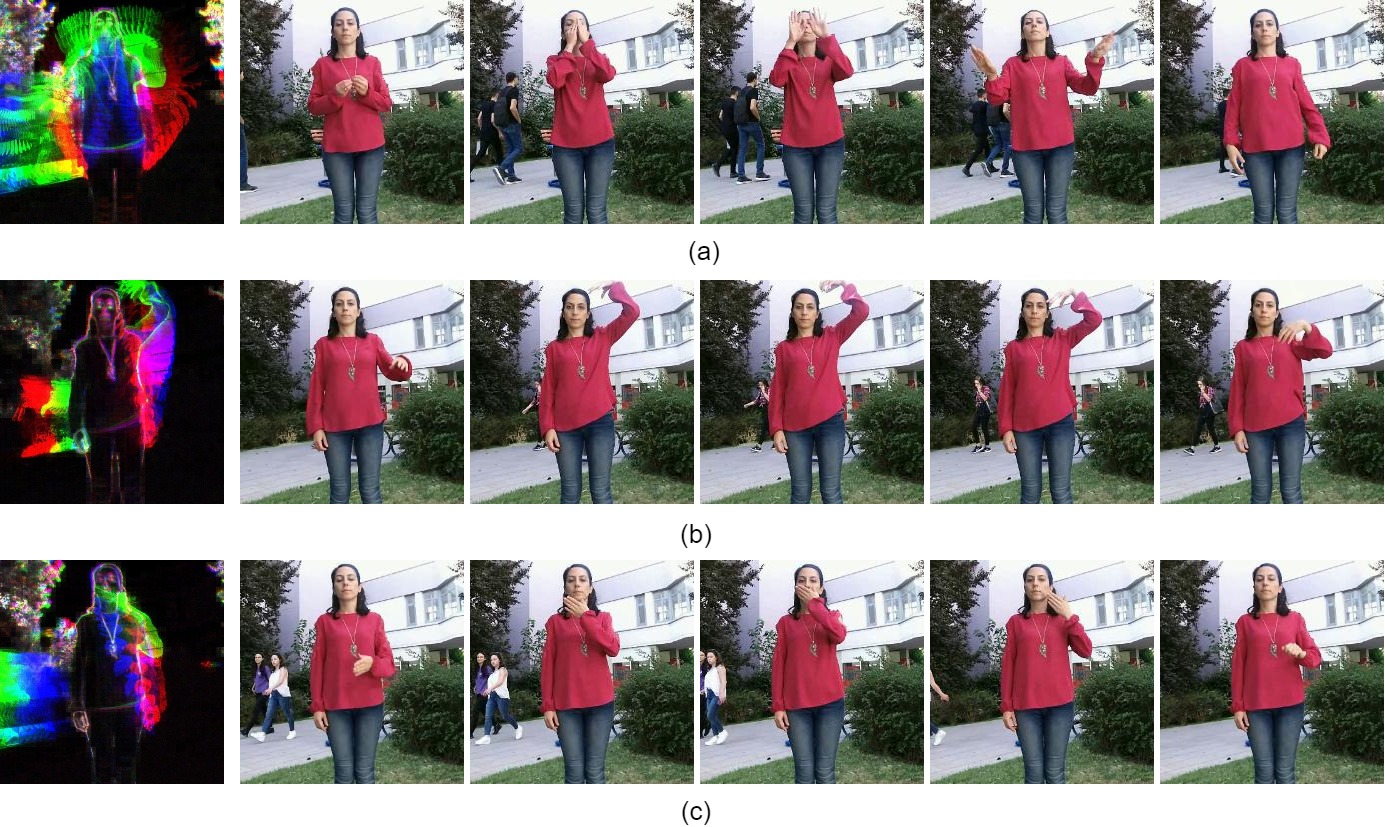}
	\caption{Sample videos that are correctly classified by the I3D, I3D with RGB-MHI attention, and I3D+RGB-MHI fusion models. Although the backgrounds contain moving people and swaying trees, all models could identify the correct sign without getting influenced by the cluttered background.}
	\label{fig:complexBackgrounds}
\end{figure}

We observe that our proposed I3D model and its RGB-MHI variants work successfully, invariant to the complex external motion arising from the background. In dynamic backgrounds, motion history images also summarize some undesirable (moving background) information (Fig. \ref{fig:complexBackgrounds}). However, our models can tolerate external motion in the background. In Fig. \ref{fig:complexBackgrounds}, we show examples that are correctly classified with all three models, i.e. I3D, I3D with RGB-MHI attention, and I3D+RGB-MHI fusion. I3D model alone already performs well in these samples, hence we did not observe additional help from RGB-MHI images in correct classifications for these samples. Still, I3D with RGB-MHI model variants also correctly classify these challenging samples, even RGB-MHI images contain background motion.

\begin{figure} 
	\centering    
	\includegraphics[width=0.48\textwidth]{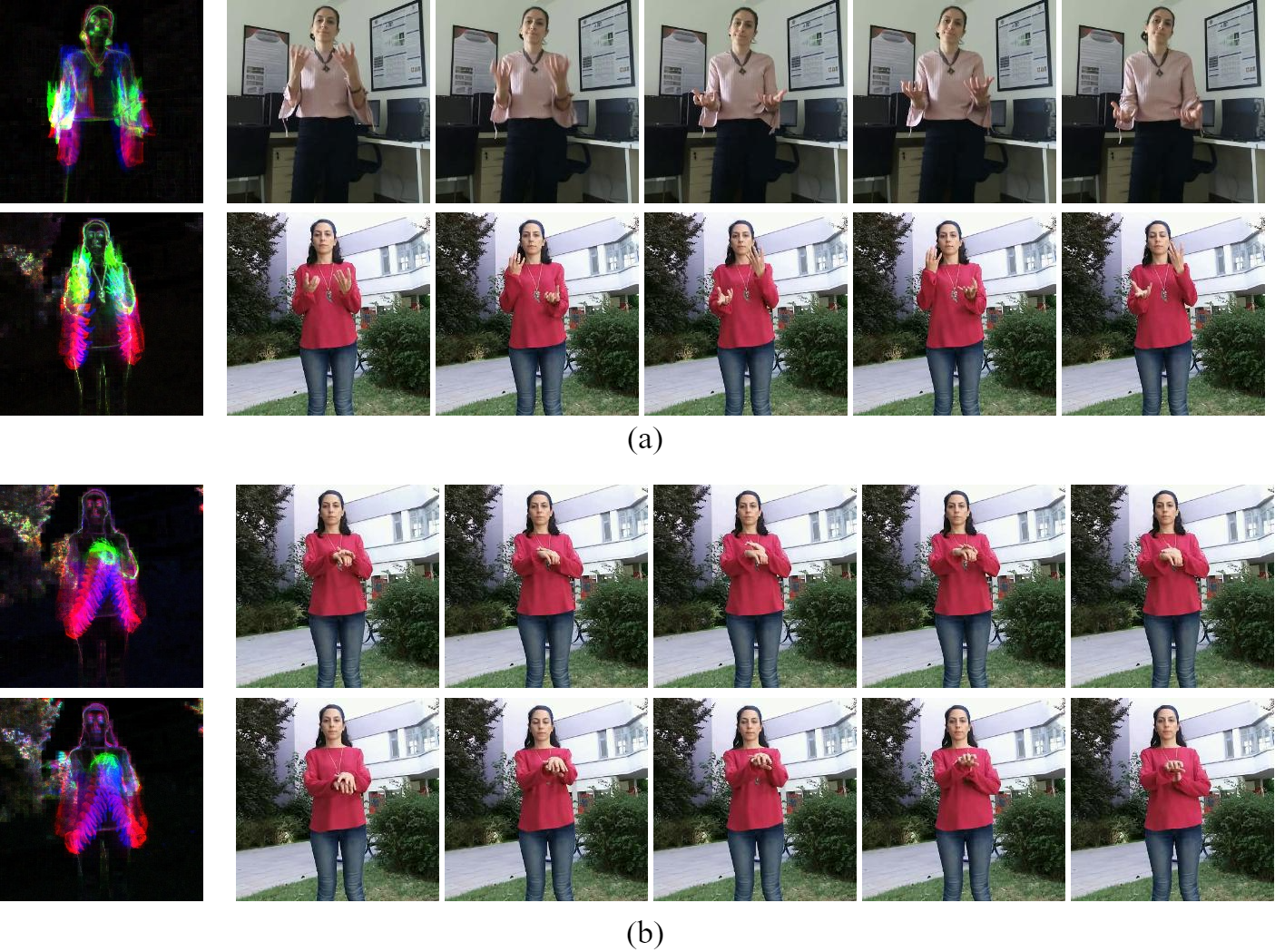}
	\caption{Sample videos that are misclassified by the I3D model and correctly classified by the I3D with RGB-MHI attention and I3D+RGB-MHI models. Each row contains some sample video frames and the corresponding RGB-MHI image. (a-first row) "agir: heavy" is misclassified with the I3D model as "ocak: oven" (second row). As seen in the video frames, hand shapes are very similar, but there is a difference in the movement of hands. In the "heavy" sign, both hands are lowered from above, while in the "oven", the hands move in opposite directions. (b-first row) "yavas: slow" sign is misclassified as "eldiven, gloves" (second row). These two signs are very similar; there is only a slight difference in hand orientation and movement.}
	\label{fig:modelComparisions}
\end{figure}

On the other hand, the overall test scores depict that the RGB-MHI model helps I3D to better classify the challenging signs in the AUTSL dataset. Note the accuracy rates of I3D, I3D with RGB-MHI attention, and I3D+RGB-MHI fusion in Table \ref{tab:our-model}, which are 89.30\%, 90.18\%, 91.13\%, respectively. When we analyze test results, we observe that while some videos are misclassified with I3D, they are classified correctly when the RGB-MHI features are incorporated. We observe that using RGB-MHI features helps to correctly classify some challenging signs where the hand shape is quite similar, yet there are differences in hand movements (Fig. \ref{fig:modelComparisions}). Note that in these test samples, the background environments are not simple.

\textbf{I3D with RGB-MHI Attention and Fusion Model:} Finally, we investigated the contribution of both the RGB-MHI attention and RGB-MHI model fusion. When we fuse the I3D with RGB-MHI attention model and the RGB-MHI model, we obtain 91.55\%. Considering the trade-off between the model complexity and accuracy rates, we selected the I3D+RGB-MHI fusion model as our final model.

In order to increase the generalization capability of the model, we train our base I3D model with the augmented data that is described in Section \ref{sec:datasets}. In this experiment, the classification accuracy has been improved from 89.30\% to 92.19\% when 6 times more train data, which is generated with data augmentation, has been used. In an SLR challenge \cite{sincan2021chalearn} on the AUTSL dataset, participants are allowed to use validation labels to fine-tune their models. Therefore, we fine-tune our model with the union of training and validation sets and obtain 92.43\% classification accuracy. In this experiment, the initial learning rate is set to $2e-3$, instead of $1e-2$. Since there is no validation set in this experiment, we stop fine-tuning when the training loss is reduced to the same level as in the previous run. Finally, we ensemble it with the RGB-MHI model using a predefined set of fusion weights. Since I3D achieves a higher accuracy rate than the RGB-MHI model, we give higher or equal weights to I3D model and we experiment with three different weight pairs for fusion: (0.7, 0.3), (0.6, 0.4), (0.5, 0.5), respectively. The fusion model achieves approximately 1\% better, and our best I3D+RGB-MHI model performs with 93.53\% accuracy, using (0.6, 0.4) weight pairs.

Table \ref{tab:comparisionAUTSL} contains the state-of-the-art model performances on AUTSL in the literature. All recent models perform considerably higher than the baseline 2D-CNN+LSTM based model. Most of the works use multiple modalities, such as skeleton joints, optical flow, hand crops, etc., and model ensembles to increase the classification accuracies. Our model performs comparable to the state-of-the-art models, although we only used RGB frames and the RGB-MHI image that we generate from RGB data. 

\begin{table*}[htbp]
	\caption{Comparison of the literature on AUTSL.}
	\centering
    \setlength\tabcolsep{2.0pt}
    \small
    \renewcommand{\arraystretch}{1.2} 
	\label{tab:comparisionAUTSL}
	\begin{tabular}{ lp{6.5cm}c }
		\hline
		\textbf{Method} & \textbf{Modalities} &  \textbf{Test Acc (\%)} \\ \hline
		SAM-SLR\cite{jiang2021skeleton}  & RGB, skeleton joints, bones, joint motion, bone motion, skeleton features, optical flow  & 98.42 \\
		MS-G3D \cite{vazquez2021isolated} & RGB, skeleton joints, bones, joint motion, bone motion & 96.15 \\ 
		I3D-VLE-transformer \cite{gruber2021mutual} & RGB, hand crops, pose, location vectors   & 95.46 \\ 
		\textbf{I3D+RGB-MHI}  & \textbf{RGB, RGB-MHI image} & \textbf{93.53} \\ 
		VTN-PF \cite{de2021isolated}  & RGB, hand crops, location vectors, skeleton  & 92.92 \\ 
		OpenPose+Holistic \cite{moryossef2021evaluating}  & Skeleton  & 81.93 \\ 
		2DCNN+Attentioned BLSTM \cite{sincan2020autsl}  & RGB  & 49.22 \\ \hline

	\end{tabular}
\end{table*}

\subsection{Experiments on BosphorusSign22k}
\label{sec:experimentsBosp}

We evaluate our I3D+RGB-MHI fusion model on another large-scale sign language dataset, BosphorusSign22k \cite{ozdemir2020bosphorussign22k}, since our I3D+RGB-MHI fusion model achieves a better result than I3D with our RGB-MHI attention model. RGB-MHI model alone classifies the signs in this dataset with 75.77\% accuracy. Following that, we conduct similar experiments with the I3D model using augmented train data. In our first experiment, we use the BSL-1K pre-trained version of the I3D model and obtained 91.64\% accuracy. Next, in order to analyze the results of the transfer learning from different datasets, we use AUTSL-trained parameters instead of BSL-1K in the same model. In this case, we obtain slightly better accuracy, i.e. 92.66\%. We first thought that pretraining with AUTSL is more helpful since both datasets are TSL datasets, and hence contain similar signs (although performed in different contexts and with different signers). However, we realized from the works of \cite{vazquez2021isolated} that pretraining with AUTSL is effective in Spanish Sign Language as well. In Vazquez-Enriquez \emph{et al.}'s work \cite{vazquez2021isolated}, the authors show the effect of transferring data from different sign language datasets. In their experiments, they worked on LSE-Lex 40 \cite{docio2020lse_uvigo} Spanish Sign Language dataset; they trained their model from scratch or applied transfer learning to models that were trained with AUTSL or WLASL datasets. They observed that the best results are obtained with an AUTSL pre-trained model. In experiments conducted on both BosphorusSign22k and LSE-Lex 40 datasets, higher accuracy rates are obtained when initial parameters are transferred from the AUTSL dataset, indicating that AUTSL provides a good initialisation for other sign language datasets. Therefore, we choose the AUTSL pre-trained I3D model as our baseline in these experiments. Then, we fine-tune the I3D model with the union of the train and validation datasets. Since there are only 4 signers in the training data, an additional 1 person in the validation set contributed to the increase in generalization of the model, and we obtained 94.47\% accuracy. Finally, we fuse our I3D and RGB-MHI model with the coefficients of $w\textsubscript{1}=0.6$; $w\textsubscript{2}=0.4$, and obtained 94.83\% accuracy. Our results are provided in Table \ref{tab:bosp_results}.

Table \ref{tab:comparisionBosp} compares model performances on BosphorusSign22k in the literature. As seen in the table, our proposed I3D+RGB-MHI fusion model achieves competitive performance, 94.83\%, with the state-of-the-art; following the best performing method very closely. Notice that in our proposed model, no explicit part segmentation is included as a separate modality. The context provided with the RGB-MHI image helps the 3D model to focus on relevant parts of the sign action.

\begin{table*}[htbp]
	\caption{Performances of the proposed methods on BosphorusSign22k.}
	\centering
    \small
	\label{tab:bosp_results}
	\renewcommand{\arraystretch}{1.2} 
	\begin{tabular}{ lcc }
		\hline
		\textbf{Model} & \textbf{Pretrained Dataset} & \textbf{Test Acc (\%)} \\ \hline
		RGB-MHI & ImageNet & 75.77 \\
		I3D & BSL-1K &  91.64 \\
		I3D &  AUTSL &  92.66 \\ 
		I3D fine tune with train+validation data & AUTSL &  94.47 \\ 
		I3D+RGB-MHI fusion & AUTSL, ImageNet & \textbf{94.83} \\  \hline
	\end{tabular}
\end{table*}

\begin{table*}[htbp]
	\caption{Comparison of the literature on BosphorusSign22k.}
	\centering
    \setlength\tabcolsep{2.0pt}
    \renewcommand{\arraystretch}{1.2} 
    \small
	\label{tab:comparisionBosp}
	\begin{tabular}{ lp{6cm}c }
		\hline
		\textbf{Method} & \textbf{Modalities} &  \textbf{Test Acc (\%)} \\ \hline
		MC3-18 Spatio-Temporal Sampling \cite{gokce2020score} & RGB, cropped hands, cropped face & 94.94 \\ 
		\textbf{I3D+RGB-MHI } & \textbf{RGB, RGB-MHI image}   & \textbf{94.83} \\ 
		IDT \cite{ozdemir2020bosphorussign22k}  & HOG, HOF, MBH & 88.53 \\ 
		MC3-18 Spatio-Temporal Sampling \cite{gokce2020score} & RGB & 86.91 \\ 
		MC3-18 \cite{ozdemir2020bosphorussign22k} & RGB & 78.85 \\ \hline

	\end{tabular}
\end{table*}

\section{Conclusion}
\label{sec:conclusion}
In the isolated SLR literature, the methods that train different models with multiple modalities and use ensembles of these models with various fusion techniques achieve more successful results. Although performing significantly well, such models are complex and usually demand more computational resources. In this research, our aim is to utilize only the RGB data modality to avoid additional modalities without sacrificing the model performances significantly. Two methods are proposed in this regard; in the first approach, rather than explicitly segmenting the hands or face, a model that focuses on motion patterns is proposed with the RGB-MHI images. For the second approach, RGB and RGB-MHI features are fused in a late fusion technique, before the prediction. Compared to the literature, our proposed models perform comparably to the state-of-the-art models that use multiple data modalities; and provide strong baselines for the models that use RGB data on two large-scale TSL benchmarks. As future work, we plan to investigate the effectiveness of the proposed models in the continuous SLR domain.

\section*{Acknowledgment}

This research is part of a project funded by TUBITAK (The Scientific and Technological Research Council of Turkey) under the grant number 217E022. The numerical calculations reported in this paper were partially performed at TUBITAK ULAKBIM, High Performance and Grid Computing Center (TRUBA resources). Authors thank TUBITAK for the support. 

\bibliographystyle{ieeetr}
\footnotesize \bibliography{main}

\end{document}